\documentclass[journal]{IEEEtran}
\usepackage{cite}
\usepackage{amsmath,amssymb,amsfonts}
\usepackage{url}
\usepackage{amsthm}
\usepackage{algorithmic}
\usepackage{mathrsfs}
\usepackage{graphicx}
\usepackage{textcomp}
\usepackage{algorithm}
\usepackage{booktabs}
\usepackage{tabularx}
\usepackage{multirow}
\usepackage{arydshln}
\usepackage{array}
\usepackage{utfsym}
\usepackage{ragged2e}
\usepackage{adjustbox}
\usepackage{hyperref}
\usepackage{xcolor}
\usepackage{booktabs}
\usepackage{multirow}
\usepackage{graphicx}
\usepackage{tikz}
\usepackage{amssymb}
\usepackage{pifont}
\usepackage{wasysym}
\usepackage{color}
\usepackage{colortbl}  %%1111
\usepackage{pifont}
\newcommand{\Checkmark}{\ding{51}}
\usepackage{amsfonts, mathrsfs}
\usepackage{float}
\usepackage{pifont} % for star symbol
\newcommand{\mystar}{{\Pisymbol{pzd}{72}}}
\begin{document}
\title{Serp-Mamba: Advancing High-Resolution \\Retinal Vessel Segmentation with Selective State-Space Model
}
\author{Hongqiu Wang, Yixian Chen, Wu Chen, Huihui Xu, Haoyu Zhao, Bin Sheng, Huazhu Fu, Guang Yang, Lei Zhu
\vspace{-6mm}
% \thanks{This work was supported by the ... Corresponding author: Lei Zhu (Leizhu@ust.hk).}
\thanks{H. Wang and H. Xu are with the Department of Systems Hub, Hong Kong University of Science and Technology (Guangzhou), Guangzhou 511400, China (e-mail: hwang007@connect.hkust-gz.edu.cn).}
\thanks{Y. Chen and W. Chen are with the School of Information and Communication Engineering, University of Electronic Science and Technology of China, Chengdu 610072, China.}
\thanks{H. Zhao is with the School of Computer Science, Wuhan University, Wuhan, 430000, China.}
\thanks{Bin Sheng is with the Department of Computer Science and Engineering, Shanghai Jiao Tong University, Shanghai 200240, China.}
\thanks{Huazhu Fu is with the Institute of High Performance Computing (IHPC), Agency for Science, Technology and Research (A*STAR), 138632, Singapore 
(e-mail: hzfu@ieee.org).}
\thanks{Guang Yang is with Bioengineering/Imperial-X, Imperial College London, UK (e-mail: g.yang@imperial.ac.uk).}
\thanks{Lei Zhu is with Systems Hub, Hong Kong University of Science and Technology (Guangzhou), China, and The Hong Kong University of Science and Technology, Hong Kong SAR, China  (E-mail: leizhu@ust.hk).}
\thanks{Lei Zhu (leizhu@ust.hk) is the corresponding author of this work.}
}

\maketitle
\begin{abstract}
%%背景介绍
Ultra-Wide-Field Scanning Laser Ophthalmoscopy (UWF-SLO) images capture high-resolution views of the retina with typically 200 spanning degrees. Accurate segmentation of vessels in UWF-SLO images is essential for detecting and diagnosing fundus disease.
%%方法背景介绍
Recent studies have revealed that the selective State Space Model (SSM) in Mamba performs well in modeling long-range dependencies, which is crucial for capturing the continuity of elongated vessel structures. Inspired by this, we propose the Serpentine Mamba (Serp-Mamba) network to address this challenging task.
Specifically, we recognize the intricate, varied, and delicate nature of the tubular structure of vessels. Furthermore, the high-resolution of UWF-SLO images exacerbates the imbalance between the vessel and background categories.
%%两种方法介绍
Based on the above observations, we first devise a Serpentine Interwoven Adaptive (SIA) scan mechanism, which scans UWF-SLO images along curved vessel structures in a snake-like crawling manner. This approach, consistent with vascular texture transformations, ensures the effective and continuous capture of curved vascular structure features.
Second, we propose an Ambiguity-Driven Dual Recalibration (ADDR) module to address the category imbalance problem intensified by high-resolution images. Our ADDR module delineates pixels by two learnable thresholds and refines ambiguous pixels through a dual-driven strategy, thereby accurately distinguishing vessels and background regions. 
Experiment results on three datasets demonstrate the superior performance of our Serp-Mamba on high-resolution vessel segmentation. We also conduct a series of ablation studies to verify the impact of our designs. Our code shall be released upon publication (\href{https://github.com/whq-xxh/Serp-Mamba}{\textit{Git}}).
\end{abstract}

\section{Introduction}
\label{sec:introduction}
%%UWFSLO重要
\IEEEPARstart{U}{ltra-wide-field} Scanning Laser Ophthalmoscopy (UWF-SLO) imaging represents a significant advancement in retinal imaging technology, providing a wide 200 degrees Field-of-View (FOV) in a single high-resolution image \cite{ding2020weakly,tang2024applications}. These detailed images are crucial in medical image analysis, especially for subsequent tasks like disease diagnosis, monitoring, and treatment planning. By offering a comprehensive view of the retinal periphery, UWF-SLO images enable the detection of peripheral retinal abnormalities that may be missed by conventional imaging techniques \cite{ju2021leveraging,ding2020weakly}.

\begin{figure}[t]
    \centering
    \includegraphics[width=0.5\textwidth]{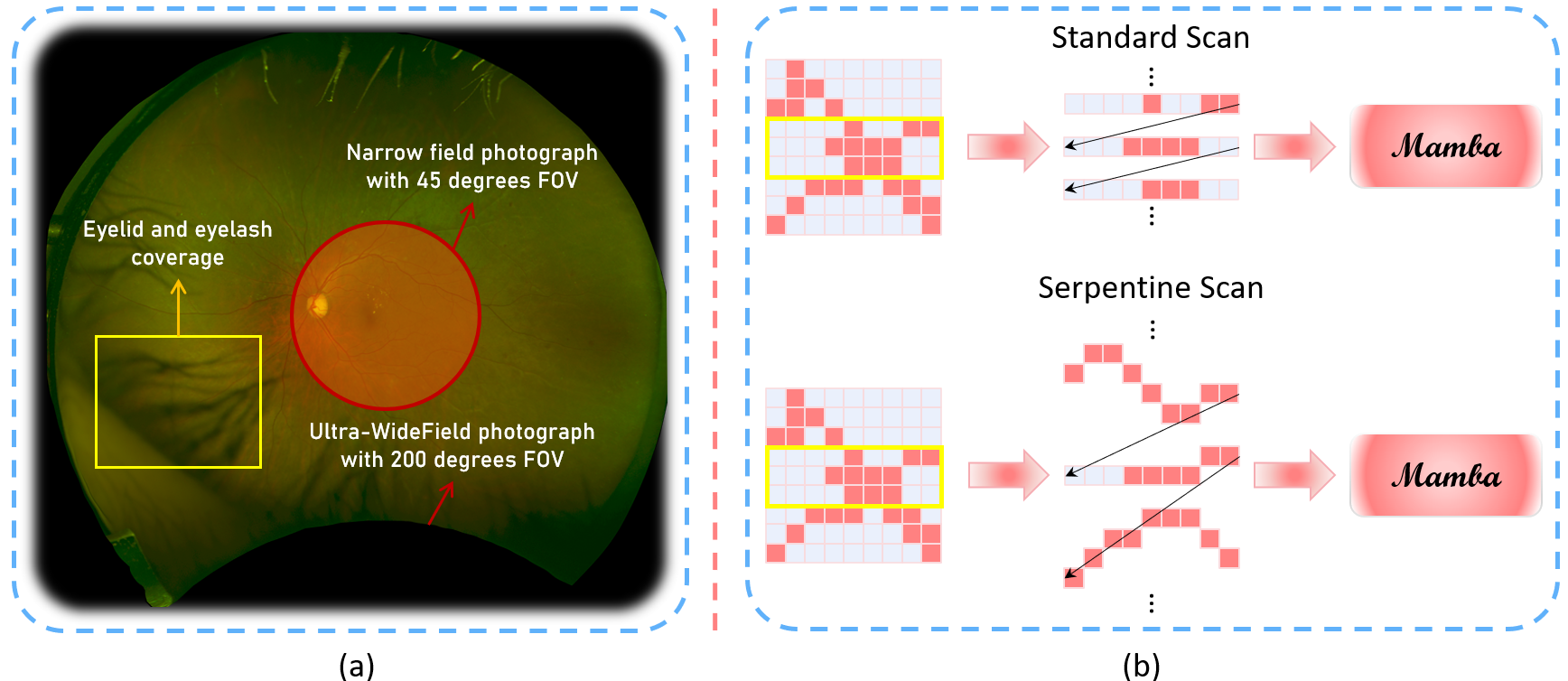}
    \vspace{-7mm}
    \caption{(a) Comparison of FOV between ultra-widefield photograph and narrow field photograph; (b) Comparison of fixed scan path and our learnable serpentine scan path for Mamba Block.}
    \label{fig:comparison}
    \vspace{-5mm}
\end{figure}

%%UWFSLO难 
Although UWF-SLO images provide high resolution and a wide FOV, limited studies are based on UWF-SLO. Meanwhile, existing studies still leave significant room for improvement, as many methods achieve segmentation Dice scores of less than 60\% \cite{qiu2023rethinking}. This is because UWF-SLO adds increased complexities and challenges to the segmentation of retinal vessels \cite{nagiel2016ultra,silva2012nonmydriatic}. The main reasons include the following:
Firstly, shown in Fig.~\ref{fig:comparison} (a), UWF-SLO images cover a wide area of the fundus, especially the eyelids, and eyelashes, which may cover part of the blood vessels, increasing the segmentation difficulty \cite{ding2022combining,ju2021leveraging}. Secondly, the high-resolution UWF-SLO images show uneven illumination and contrast between blood vessels and background which can complicate the segmentation process, leading to inaccuracies \cite{ding2020weakly}.
Thirdly, vessels occupy a very small percentage of the entire fundus image, less than a few percent (only 2.7\% in Fig.~\ref{fig:comparison} (a)), resulting in a significant class imbalance. This makes distinguishing blood vessels from the background exceedingly challenging.
These challenges necessitate the development of robust and efficient image segmenting algorithms to fully unleash the potential of UWF-SLO imaging in clinical applications.

%%mamba模型
Recently, Mamba as a State Space Model (SSM) has shown strong capabilities in modeling long-range dependencies \cite{gu2023mamba}. There are more and more efforts introducing Mamba into different research fields \cite{heidari2024computation,zhang2024survey}, such as medical image segmentation \cite{xing2024segmamba,yang2024vivim,liao2024lightm}, remote sensing image segmentation \cite{ma2024rs,chen2024rsmamba}, video understanding \cite{chen2024video}, image restoration \cite{guo2024mambair}. For example, Mamba-UNet adopts an encoder-decoder structure purely based on visual Mamba and incorporates skip connections to preserve spatial information \cite{wang2402mamba}. VM-UNet introduces the visual state space (VSS) block as a basic block to capture extensive contextual information and constructs an asymmetric encoder-decoder structure \cite{ruan2024vm}.

% Nonetheless, there is still significant potential within Mamba, especially for UWF-SLO vessels, the high-resolution tubular structure segmentation. Currently, there are plenty of modifications for Mamba, and most of the improvements are made by adding more scanning directions to solve the orientation sensitivity problem faced by applying Mamba to the vision field \cite{liu2024vmamba, zhao2024rs}. However, none of these related methods have made dedicated improvements to Mamba for fundus vessel segmentation.
% %%
% Meanwhile, despite Mamba's robust capability in long-sequence modeling, it faces challenges in maintaining continuity along curved and variable vessel structures and the intensified category imbalance in high-resolution images. In addressing the challenge of blood vessel segmentation in UWF-SLO images, dedicated modifications for the original Mamba are necessary.

Nonetheless, current SSM-based methods lack specific adaptations for vessel (tubular structure) segmentation. Most improvements to the original Mamba focus on adding fixed, non-learnable scanning directions to address orientation sensitivity issues in vision tasks \cite{liu2024vmamba, zhao2024rs}. However, these methods are unable to adapt to the curved and variable nature of blood vessels, failing to maintain vessel continuity. To effectively segment vessels in high-resolution UWF-SLO images, dedicated modifications to Mamba are necessary.

In this paper, we take the \textit{first} step by proposing the Serp-Mamba for vessel segmentation of high-resolution UWF-SLO images, aiming to address the challenges mentioned above. This architecture is particularly designed for the tubular and curved structure, thus allowing the model to focus more on the vessel features.
Firstly, we devise a more specialized scanning strategy, named the Serpentine Interwoven Adaptive (SIA) scan mechanism, where the Mamba block no longer scans in fixed directions but scans along the learned deformable trails, mimicking the snake's adaptive crawling behavior. This allows Mamba to overcome the problem of orientation sensitivity while ensuring the effective extraction of continuous vascular features in high-resolution images.
Furthermore, we propose the Ambiguity-Driven Dual Recalibration (ADDR) module. Recognizing the prior knowledge of the topological consistency of blood vessels \cite{zhao2019retinal}, we first develop a Continuity Perception mechanism to detect the pixel values around the ambiguous pixels, thereby ensuring the continuity of blood vessels.
Subsequently, ADDR will refine the ambiguous pixels through the dual-driven of blood vessels and background. By integrating the Continuity Perception with the dual-driven approach, the ADDR module effectively addresses the class imbalance problem, while also guaranteeing that the segmented vessels are effectively attended to and maintain coherence (examples in Fig.~\ref{fig:ADDRview}).
Experiments on three datasets demonstrate that the Serp-Mamba outperforms the state-of-the-art methods on several key metrics. We also demonstrate the effectiveness of our designs through ablation experiments. The main contributions could be summarized as follows: 

\begin{itemize}
\item To the best of our knowledge, we are the first Mamba-based work for high-resolution vessel segmentation. The proposed Serp-Mamba offers an innovative solution, adeptly tackling tubular continuity and class imbalance challenges in vessel segmentation within UWF-SLO.
\item We devise a novel Serpentine Interwoven Adaptive scan mechanism that enables Mamba to scan tortuous blood vessels in a snake-like crawling manner, accurately capturing vascular features and ensuring continuity.
\item We propose an Ambiguity-Driven Dual Recalibration module, which refines the ambiguous pixel through the dual-driven of vessels and background, thus addressing the category imbalance problem effectively.
\item Experiments on three UWF-SLO datasets demonstrate that our method outperforms other vessel, super-resolution and Mamba-based segmentation methods. Moreover, ablation experiments reveal the effectiveness of our designs.
\end{itemize}

\begin{figure*}[t]
    \centering
    \includegraphics[width=1\textwidth]{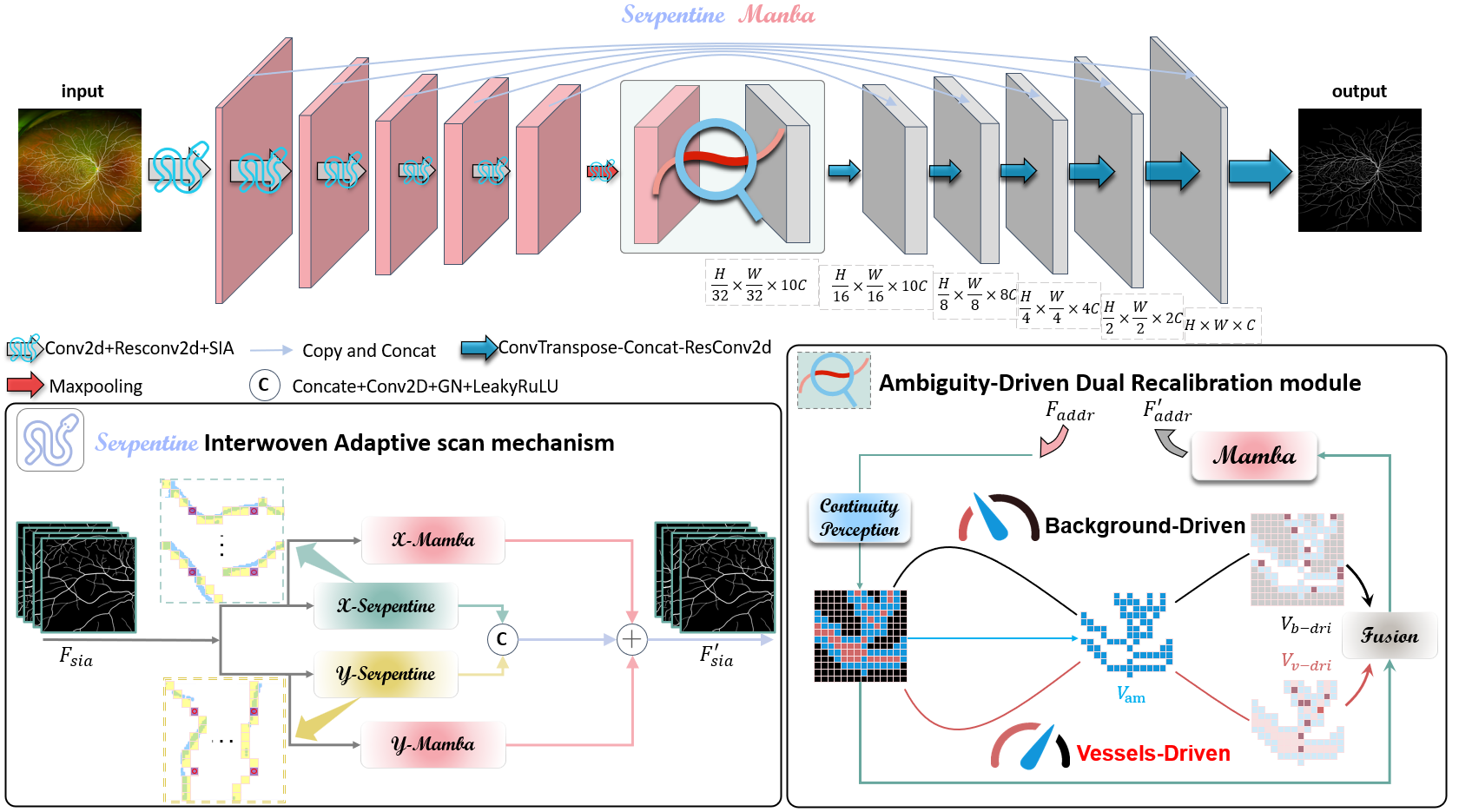}
    \vspace{-7mm}
    \caption{Overview of the Serp-Mamba. The top of the figure shows the whole model. Our proposed method adopts a classical U-shaped structure which includes the novel Serpentine Interwoven Adaptive (SIA) scan mechanism and the Ambiguity-Driven Dual Recalibration (ADDR) module. As shown in the lower-left corner of the figure, in SIA, the input features will be scanned by the Serpentine Mamba scan and the ordinary Mamba scan in the $\mathcal{X}$ and $\mathcal{Y}$ directions respectively, and then combined. For details of the SIA mechanism, please refer to Fig.~\ref{fig: explain of SIA}. In the ADDR module in the lower right corner, the pixels of the input feature are first distinguished by two learnable thresholds, and then the ambiguous pixels are refined through continuity perception and dual driving of blood vessels and background pixels. For details of the ADDR module, please refer to Fig.~\ref{fig: addr}.}
    \label{fig:Serpentine Mamba}
    \vspace{-3mm}
\end{figure*}

\section{Related work}
\subsection{Retinal Vessel Segmentation}
Retinal vessel segmentation is an essential field in medical image processing, aimed at delineating vascular networks from retinal images.
By analyzing blood vessel morphology and changes, clinicians can diagnose ophthalmic diseases such as diabetes and hypertensive retinopathy \cite{hanssen2022retinal}.
Achieving precise segmentation of blood vessels presents a significant challenge due to the minuscule proportion of blood vessels within the image, coupled with their complexity and variability.

Numerous studies have developed sophisticated models to achieve more detailed and comprehensive vascular segmentation.
For example, Ryu \textit{et al.} \cite{ryu2023segr} introduce SegR-Net, which combines feature extraction, deep feature amplification, and dense multi-scale fusion to enhance the accuracy of segmentation masks.
Zhou \textit{et al.} \cite{zhou2023dual} introduce MCDAU-Net, which segments fundus images into concentric patches for detailed vessel extraction and uses a combined cascaded sparse spatial pyramid pooling and InceptionConv to reduce background noise.
Lin \textit{et al.} \cite{lin2023stimulus} propose SGAT-Net, a retinal vessel segmentation framework using adaptive modules to boost feature extraction and context awareness.
Liu \textit{et al.} \cite{liu2023aa} introduce an Attention-Augmented Wasserstein Generative Adversarial Network for this task. The network effectively manages complex vascular structures by emphasizing critical areas and capturing pixel correlations.
However, traditional fundus images are limited to the retina's central region. The introduction of UWF-SLO provides significant clinical advantages by broadening the observable area.

UWF-SLO, an advanced imaging technique, captures a broader retinal area than standard fundus photography, enhancing the diagnosis and monitoring of retinal diseases like diabetic retinopathy and age-related macular degeneration \cite{tang2024applications}. However, the unique characteristics of UWF-SLO images pose special challenges for accurate vessel segmentation.
Recently, Shin \textit{et al.} \cite{shin2019deep} propose a Vessel Graph Network (VGN) by combining GNN with CNN to leverage the continuity relationship between blood vessels and thus improve the segmentation accuracy.
Ding \textit{et al.} \cite{ding2020weakly} propose an annotation-efficient framework for vessel detection in UWF-SLO photography without the need to annotate UWF-SLO vessel maps from scratch.
Li \textit{et al.} \cite{li2023minet} develop a multi-input fusion module that captures multi-scale features and is further refined by a multiscale spatial pyramid for enhanced segmentation results.
Qiu \textit{et al.} \cite{qiu2023rethinking} rethink the dual-stream learning framework and propose a DS2F (Dual-Stream Shared Feature) framework, including a shared feature extraction module.
Kim \textit{et al.} \cite{kim2024c} introduce the C-DARL model, a self-supervised vessel segmentation method that integrates diffusion and generation modules with contrastive learning to improve performance.
However, these methods may struggle to consistently capture the continuity features of the vessels, which can occasionally lead to broken segments. In our approach, the Serp-Mamba focuses on the vessel features to complete the continued scan and avoids the model being affected by the large background area through ADDR, which effectively overcomes multiple challenges posed by UWF-SLO images.

\begin{figure*}[t]
    \centering
    \includegraphics[width=1\textwidth]{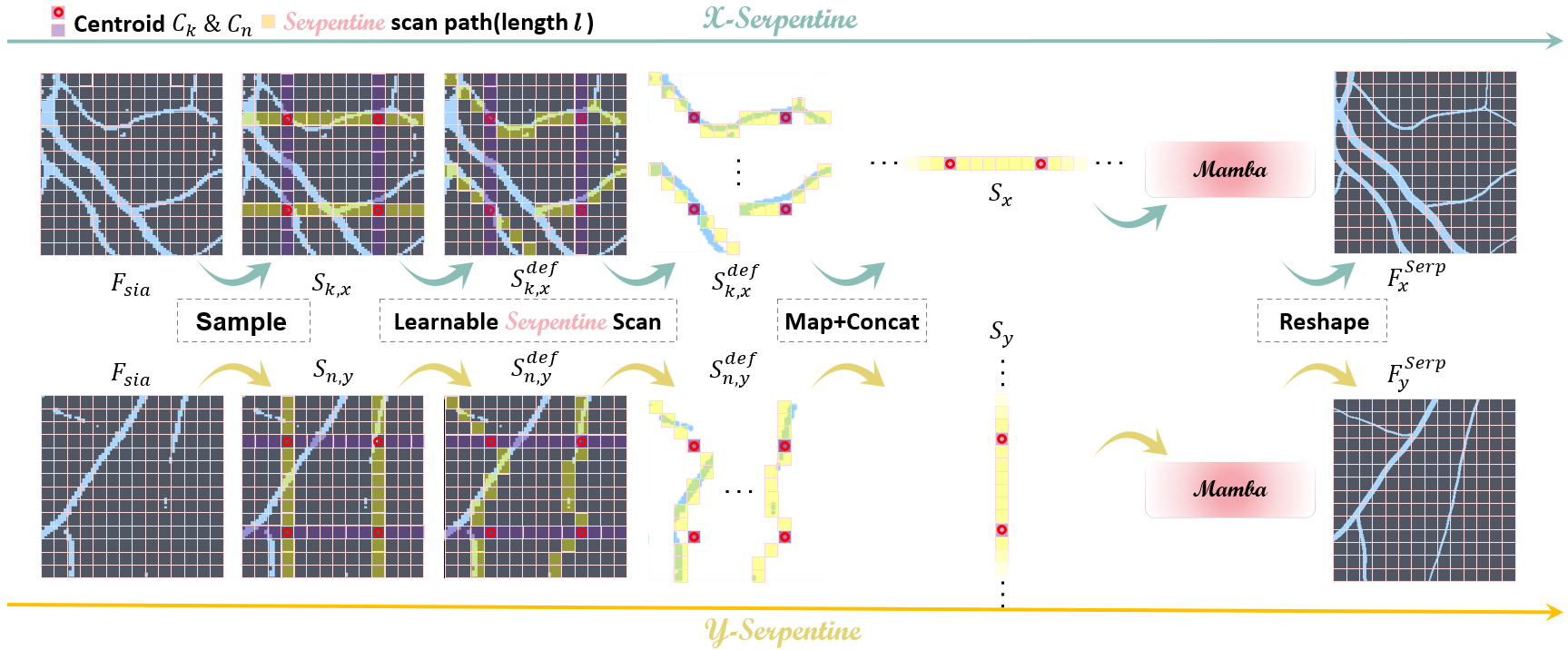}
    \vspace{-7mm}
    \caption{Illustration of the proposed serpentine scan mechanism along the $\mathcal{X}$ and $\mathcal{Y}$ axes. The upper and lower left of the figure shows two patches of $F_{sia}$.
    i) Take the $\mathcal{X}$ direction (the first row) for instance, we set the serpentine scan path length $l$ ($l$=7 here). We sample pixels at fixed intervals along the $\mathcal{X}$ direction as center points (purple boxes).
    ii) The learnable serpentine scan modifies these paths (yellow boxes) through deformation learning based on these center points. 
    iii) All serpentine scan paths are mapped to straight lines and concatenated into a horizontal sequence $S_{x}$, which is then fed into the Mamba block to learn deep contextual information. 
    iv) Finally, the output sequence is reshaped back to the original input feature map size to obtain the final output. 
    The same process applies in the $\mathcal{Y}$ direction (the second row).}
    \label{fig: explain of SIA}
    \vspace{-4mm}
\end{figure*}

\subsection{Mamba for Medical Image Segmentation}
Medical image segmentation has always been widely studied \cite{tian2023delineation,wang2024dual,wang2024video,zhao2024morestyle}. Recently, The Mamba network has become a focal point of research \cite{gu2023mamba}. It is designed to model long-range dependencies and to improve training efficiency \cite{zhang2024survey}. 
%%Umamba, U型结构引入
To explore the potential of using Mamba blocks for medical image segmentation, U-Mamba \cite{ma2024u} integrates the SSM with the traditional UNet model, which improves the accuracy of segmentation and robustness through the long-range dependence across multiple scales. U-Mamba also integrates deep supervision to expedite training and enhance convergence, ensuring efficiency and reliability.
%%VM-UNet，纯SSM探索
VM-UNet \cite{ruan2024vm} uses the scan expanding operation and scan merging operation to scan the input image in a total of 4 fixed directions: horizontally, vertically, and their inversions. This expands Mamba's ability to capture global visual contextual information.
%% RS-Mamba，更多基本扫描方向
Meanwhile, there are also studies exploring more basic scanning paths for Mamba. RS-Mamba \cite{zhao2024rsmambalargeremotesensing} proposes an omnidirectional selective scanning (OSS) module, which can extract large-scale spatial features in multiple directions.
Despite the significant contributions made by the studies mentioned above, there are limitations when processing UWF-SLO vessel segmentation.
Firstly, as the tubular structure, blood vessels are curved and variable, while the above methods only scan in fixed directions. Secondly, the high resolution of UWF-SLO images causes extreme class imbalance problems.
In our work, we propose the SIA scan mechanisms and ADDR module, enabling the Serp-Mamba to dynamically adapt its scanning path for continuous tracking of curved vessel features and effectively tackle class imbalance in high-resolution vascular segmentation.

\section{Methodology}
In Fig.~\ref{fig:Serpentine Mamba}, we show the overall framework of Serp-Mamba, including the Serpentine Interwoven Adaptive (SIA) scan mechanism and the Ambiguity-Driven Dual Recalibration (ADDR) module.
In the following content, we will introduce the SIA scan mechanism in Section~\ref{SIA}, where the Mamba Block scans along the curvature of the vessel, effectively enhancing the continuity of vessel features. We then present the ADDR module in Section~\ref{ADDR}, demonstrating how vessel and background features are accurately distinguished. Finally, we will show the implementation details and evaluation metrics in Section~\ref{IDEM}.

\subsection{Serpentine Interwoven Adaptive scan} \label{SIA}
%%SIA的设计动机
Previous research has developed various scanning techniques for the Mamba model, enabling multi-directional information capture. Although these established methods have proven effective across a range of tasks, we identify limitations in their application to retinal vessel segmentation, particularly with high-resolution images.
The blood vessel structure is complicated and delicate with fine flexure, which challenges the common scanning techniques to maintain vascular continuity. In light of these considerations, we devise the SIA scan mechanism.
%% Fig2的过程解释
As illustrated in the lower left corner of Fig.~\ref{fig:Serpentine Mamba}, to more comprehensively capture the vascular features in the high-resolution UWF-SLO image, the input features undergo both ordinary Mamba scanning and serpentine Mamba scanning in the $\mathcal{X}$ and $\mathcal{Y}$ directions.
More precisely, the results of $\mathcal{X}$-$Serpentine$ and $\mathcal{Y}$-$Serpentine$ are concatenated together and then passed through the Convolution, GroupNorm and LeakyRuLU functions, where $\mathcal{X}$-$Serpentine$ and $\mathcal{Y}$-$Serpentine$ are referred to as the Serpentine scan mechanism. Finally, the results of $\mathcal{X}$-$Mamba$ and $\mathcal{Y}$-$Mamba$ are element-wise added with above result and output as $F_{sia}^{'}$.
%% 解释Serpentine Scan大致做了什么

\begin{figure*}[t]
    \centering
    \includegraphics[width=0.9\textwidth]{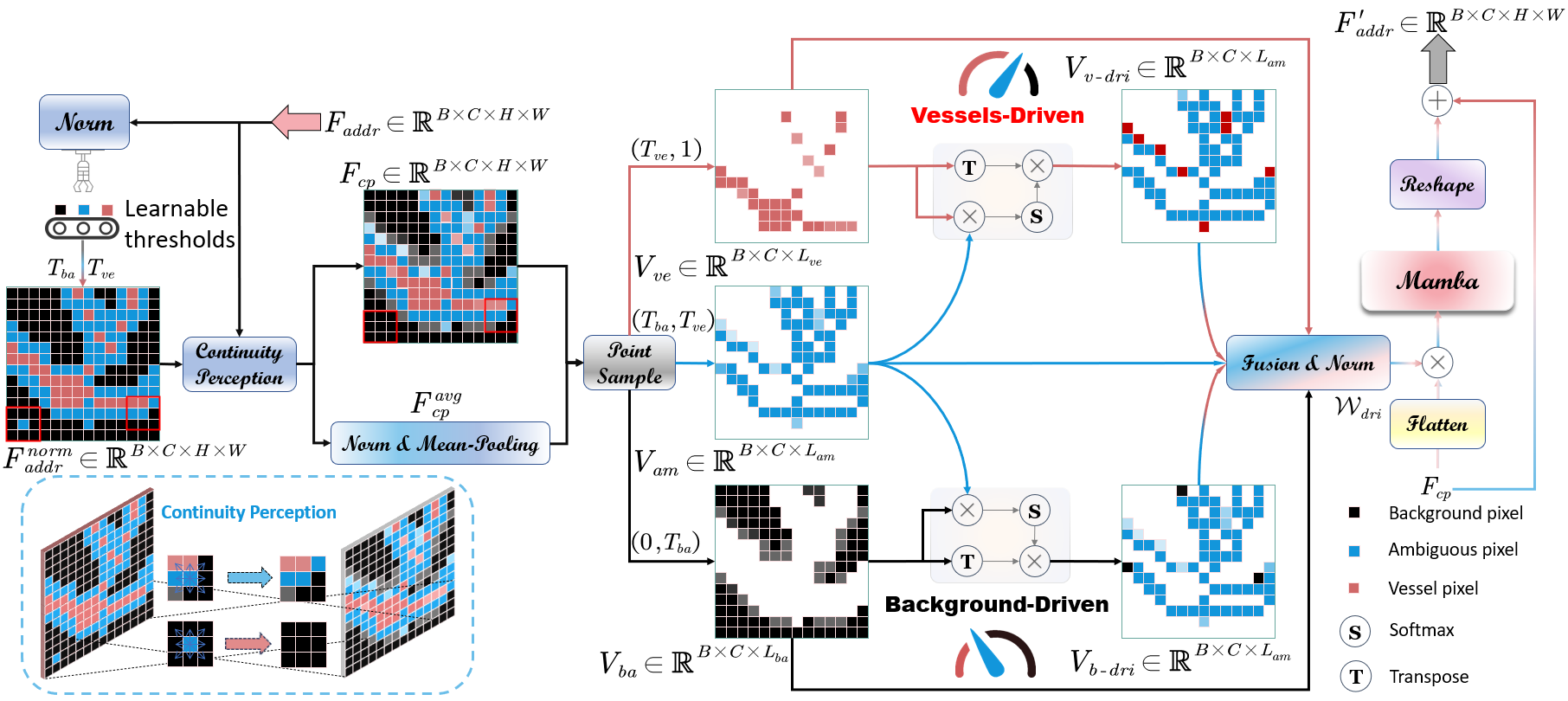}
    \vspace{-3mm}
    \caption
    {Our Ambiguity-Driven Dual Recalibration module consists of four steps: i) The two learnable Thresholds $T_{ba}$ (for background) and $T_{ve}$ (for vessels) used in the scanning process are first determined and continuously updated in subsequent training. ii) The ambiguous pixels are then processed through the Continuity Perception strategy. iii) Next, three vectors $V_{ve}, V_{am}, V_{ba}$ are obtained by applying the thresholds to distinguish vessels, ambiguous regions, and background, after which the Vessel-Driven and Background-Driven are performed. iv) Finally, the weight \(\mathcal{W}_{dri}\) is utilized to enhance results, and the Mamba block is employed to complete the scanning process. }
    \label{fig: addr}
    \vspace{-5mm}
\end{figure*}

%%Fig3 具体做法
Fig.~\ref{fig: explain of SIA} demonstrates the details of the Serpentine scan mechanism, designed to dynamically adapt the scanning path to the curvature of vessels. This allows the Mamba block to align more precisely with the orientation of vessels, thereby enhancing the capture of vascular features.
The feature map $F_{sia} \in \mathbb{R}^{B\times C\times H\times W}$ enters as the input, where $B$, $C$, $H$, and $W$ represent batch-size, channel, height, and width respectively. 
%%Then, by using Equation\ref{eq1}, we sample the feature map $F$ in the X- and Y-directions to obtain the centroids of the serpentine scan paths, and the coordinates of each centroids $C_k$ are $(x_k,y_k)$. Every two centroids are separated by $l$ points in the corresponding directions. 
We firstly sample the feature map $F_{sia}$ in the $\mathcal{X}$\mbox{-} and $\mathcal{Y}$\mbox{-} directions to obtain the centroids of the serpentine scan paths.
The following formula takes the $\mathcal{X}$\mbox{-} direction as an example of sampling:
\begin{equation}
    \begin{aligned}
        & C_{k,x} = \mathcal{S} \left( F_{sia} \right) \\
             &= \big\{ (x_k, y_k), x_k = l\lfloor \frac{k}{H} \rfloor + \frac{\left( l+1 \right)}{2}, \ \big. \\
             &\quad \big. y_k = k\ \Theta \ H , \ k \in N \big\} \\
    \end{aligned}
    \label{eq1}
\end{equation}
where $\Theta$ and $\lfloor \cdot \rfloor$ are defined as the modulo and flooring operations respectively, and $\mathcal{S}\big( \cdot \big)$ denotes the sampling function. 
$N=\left\{ 1,\cdots ,\lfloor \frac{H\cdot W}{l} \rfloor \right\}$, where $l$ is the scan path length and $\lfloor \frac{H\cdot W}{l} \rfloor$ denotes the cardinality of $C_{k,x}$.
Similarly, $C_{n,y} = \mathcal{S} \left( F_{sia}\right)$, where $\ n \in N$.
Based on these sampled centroids $C_{k,x}$ and $C_{n,y}$, the feature map $F_{sia}$ is divided into a series of sub-sequences as part of the serpentine scan paths, which are learned to deform and gradually fit the vascular path. Each sub-sequence before deformation is expressed as:
\begin{equation}
\left\{
\begin{array}{l}
    S_{k,x} = \big\{ \big( x_k + c, y_k \big) \mid c \in \{-d, \cdots, 0, \cdots, d\} \big\} \\
    S_{n,y} = \big\{ \big( x_n, y_n + c \big) \mid c \in \{-d, \cdots, 0, \cdots, d\} \big\},
\end{array}
\right.
\end{equation}
where $c \in \left\{ -d, \cdots ,0, \cdots , d\right\}$ indicates the distance from the centroid and $d = \frac{\left( l-1 \right)}{2}$. The \((x_k, y_k)\) and \((x_n, y_n)\) represent each centroid in $C_{k,x}$ and $C_{n,y}$, separately.

During the deformation of the sub-sequences $S_{k,x}$ and $S_{n,y}$, each point learns an offset $\varDelta$. This deformation process is iterative, allowing each point to continuously adjust based on the position of the previously offset point.
Furthermore, each point is capable of adjusting in both directions perpendicular to the current scan path, allowing the sub-sequence to bend with enhanced flexibility. Each deformed sub-sequence is represented as follows:
\begin{equation}
    \begin{aligned}
        S_{k,x}^{def} &= \big\{ \big( x_k - m, y_k + \sum_{i=k-m}^k \Delta y_i \big) \mid m \in \{0, \ldots, d\} \big\} \\
        &\cup \big\{ \big( x_k + m, y_k + \sum_{i=k}^{k+m} \Delta y_i \big) \mid m \in \{0, \ldots, d\} \big\}
    \end{aligned}
\end{equation}

\begin{equation}
    \begin{aligned}
        S_{n,y}^{def} &= \big\{ \big( x_n + \sum_{i=n-m}^n \Delta x_i, y_n - m \big) \mid m \in \{0, \ldots, d\} \big\} \\
        &\cup \big\{ \big( x_n + \sum_{i=n}^{n+m} \Delta x_i, y_n + m \big) \mid m \in \{0, \ldots, d\} \big\},
    \end{aligned}
\end{equation}
where the offset $\Delta y$ is generated by applying a convolution with a $l \times l$ kernel to the input $F_{sia}$. Similarly, $\Delta x$ is produced using the same convolution, but with the input transposed before the operation.
To illustrate, consider the scan in the $\mathcal{X}$\mbox{-} direction.
For $\big( x_k - m, y_k + \sum_{i=k-m}^k \Delta y_i \big)$, this represents iterating offsets $\Delta y_i$ from the center point \((x_k, y_k)\) leftward. Equally, for the right, $\big( x_k + m, y_k + \sum_{i=k}^{k+m} \Delta y_i \big)$ accumulates offsets rightward. The offsets $\Delta y_i$ indicate upward or downward shifts.

These deformed sub-sequences $S_{k,x}^{def}$ and $S_{n,y}^{def}$ will be flattened and concatenated by $\psi \big( \cdot \big)$ to obtain $S_x$ and $S_y$ for subsequent processing as follows:
\begin{equation}
\left\{ \begin{array}{l}
	S_{x}=\psi \left( \left[ S_{1,x}^{def},S_{2,x}^{def},\cdots ,S_{k,x}^{def},\cdots S_{\lfloor \frac{H\cdot W}{l} \rfloor,x}^{def} \right] \right)\\
	S_{y}=\psi \left( \left[ S_{1,y}^{def},S_{2,y}^{def},\cdots ,S_{k,y}^{def},\cdots S_{\lfloor \frac{H\cdot W}{l} \rfloor,y}^{def} \right] \right).\\
\end{array} \right. 
\label{eq4}
\end{equation}

The sequences in two directions ($S_{x}$ and $S_{y}$) are then scanned by the Mamba block. The corresponding results are expressed as $S_{x}^{\mathcal{M}}=\mathcal{M}\big( S_{x} \big)$ and $S_{y}^{\mathcal{M}}=\mathcal{M}\big( S_{y} \big)$, where $\mathcal{M}\big( \cdot \big)$ denotes Mamba feature transformation.

After the above operations, they need to be reshaped back to the feature maps $F_{x}^{Serp}$ and $F_{y}^{Serp}$ by:
\begin{equation}
\left\{ \begin{array}{l}
	F_{x}^{Serp}=\psi _{inv}\left( S_{x}^{\mathcal{M}} \right)\\
	F_{y}^{Serp}=\psi _{inv}\left( S_{y}^{\mathcal{M}} \right),\\
\end{array} \right. 
\label{eq6}
\end{equation}
where $\psi _{inv} \big( \cdot \big)$ denotes a reshape operation from a one-dimensional vector into a two-dimensional feature map. 

As illustrated in Fig.~\ref{fig:Serpentine Mamba}, to ensure comprehensive feature extraction and enhance accuracy, we first merge \( F_{x}^{Serp} \) and \( F_{y}^{Serp} \) using \( \varPhi(\cdot) \). Then, the outputs of \(\mathcal{X}\)-Mamba and \(\mathcal{Y}\)-Mamba are element-wise added to the result of \( \varPhi \big( F_x^{Serp}, F_y^{Serp} \big) \), effectively combining all components. The entire process is summarized as follows:
\begin{equation}
\begin{array}{l}
	F_{sia}^{'}=\varPhi \big( F_x^{Serp}, F_y^{Serp}\big) + \mathcal{X} \mbox{-} Mamba\left( F_{sia} \right) \\
    +\mathcal{Y} \mbox{-} Mamba\left( F_{sia} \right),\\
\end{array}
\end{equation}
where $\varPhi\big( \cdot \big)$ stands for Concatenation, Conv2D, GroupNorm, and LeakyReLU functions.

\subsection{Ambiguity-Driven Dual Recalibration} \label{ADDR}
The presence of blood vessels, which occupy only a small portion of the entire map, particularly in high-resolution UWF-SLO images, leads to an intensified category imbalance problem, impacting the segmentation accuracy. To address this, we incorporated the ADDR module into Serp-Mamba, as shown in the lower right corner of Fig.~\ref{fig:Serpentine Mamba}. This module refines visual features through a dual-driven approach, targeting both vessel and background elements to enhance segmentation accuracy and address category imbalance.

Fig.~\ref{fig: addr} depicts the specific process of our ADDR module, the input to ADDR is given as $F_{addr}$. To classify different categories of pixels, we normalize $F_{addr}$ to $\left[ 0,1 \right]$ on each channel to obtain $F_{addr}^{norm}$.
We also set two learnable thresholds $T_{ba}$ and $T_{ve}$, initialized to 0.45 and 0.55, respectively.
For pixels with values around 0.5, the model may classify them ambiguously, leading to potential misclassifications that compromise both the accuracy and continuity of the vessel segmentation.
Based on the topological characteristics of vascular continuity, we design a Continuity Perception strategy for these ambiguous pixels, as illustrated on the left side of Fig.~\ref{fig: addr}.
For each pixel in $F_{addr}^{norm}$ with a value between $T_{ba}$ and $T_{ve}$, we analyze the values of its eight neighboring pixels. Suppose all surrounding values are below $T_{ba}$, indicating background. In that case, this ambiguous pixel is also classified as background in $F_{cp}$ output, adhering to the principle of vascular continuity (isolated vessel pixels surrounded by background are adjusted for consistency). Otherwise, the current value is maintained.
The above-mentioned process can be expressed as:
%%
% \begin{equation}
%     \begin{aligned}
%         & F_c\big( i,j \big) =\big( \big[ T_{ba}\le F_{\mathcal{M}}^{norm}\big( i,j \big) \le T_{ve} \big] \cdot F_{\mathcal{M}}\big( i,j \big) \big) \cdot \omega \\
%         & +\big( \big[ T_{ba}\le F_{\mathcal{M}}^{norm}\big( i,j \big) \le T_{ve} \big] \cdot \min \big( F_{\mathcal{M}}\big( i,j \big) \big) \big) \cdot \xi 
%     \end{aligned}
%     \label{eq7}
% \end{equation}
% \begin{equation}
% \left\{ \begin{array}{l}
% 	\xi =\prod_{\big( m,n \big) \in N\big( i,j \big)}{\big[ F_{\mathcal{M}}^{norm}\big( m,n \big) <T_{ba} \big]}\\
% 	\omega =1-\xi\\
% \end{array} \right. 
% \end{equation}

\begin{algorithm}[h]
\caption{Continuity Perception strategy}
\begin{algorithmic}
\REQUIRE The pixel being judged $F_{addr}^{norm}(i,j)$, the background threshold $T_{ba}$, the vessel threshold $T_{ve}$, the set of coordinates of 8 points around the ambiguous pixel $N(i,j)$
\STATE 
\textbf{if} {$T_{ba} \leq F_{addr}^{norm}(i,j) \leq T_{ve}$}\textbf{:} 
\STATE \hspace{1em}\text{Background\_Tag}=\text{TRUE}; \text{Background\_Value}=$\min(F_{addr})$
\STATE \hspace{1em}\textbf{for} $\text{each (m,n)} \in \text{N(i,j)}$ \textbf{do}
\STATE \hspace{2em}\textbf{if} $F_{addr}^{norm}(m,n) \geq T_{ve}$\textbf{:} \text{Background\_Tag}=\text{FALSE}
\STATE \hspace{1em}\textbf{if} \text{Background\_Tag}\textbf{:} $F_{cp}(i,j)$=\text{Background\_Value}
\STATE\hspace{1em}\textbf{else:} $F_{cp}(i,j)=F_{addr}(i,j)$
\STATE \textbf{else:} $F_{cp}(i,j)=F_{addr}(i,j)$
\end{algorithmic}
% \begin{tikzpicture}[overlay, remember picture]
%     \draw[line width=0.3mm] (0.5, 0.8) -- (0.5, 3.3);
% \end{tikzpicture}
\end{algorithm}

To further process the ambiguous pixels, we need to extract the three categories of pixels in $F_{cp}$ separately. With normalization and mean-pooling (channel-dimension) operations on $F_{cp}$ to acquire $F_{cp}^{avg}$, the category interval where each pixel in $F_{cp}^{avg}$ locates is detected in sequence, and the coordinate values are retained separately to obtain the coordinate variables:
\begin{equation}
\left\{ \begin{array}{l}
	\mathcal{C}_{ba}=\left\{ \left( i,j \right) |F_{cp}^{avg}\left( i,j \right) <T_{ba} \right\}\\
	\mathcal{C}_{am}=\left\{ \left( i,j \right) |T_{ba}<F_{cp}^{avg}\left( i,j \right) <T_{ve} \right\}\\
	\mathcal{C}_{ve}=\left\{ \left( i,j \right) |F_{cp}^{avg}\left( i,j \right) >T_{ve} \right\},\\
\end{array} \right. 
\end{equation}

If $F_{cp}^{avg}(i, j) < T_{ba}$, it suggests that the pixel at coordinate (i, j) should be included in $\mathcal{C}_{ba}$, as its value falls below the threshold $T_{ba}$.
Following that, the vessel, ambiguous, and background vectors $V_{ve}, V_{am}, V_{ba}$ are point sampled from $F_{cp}$ using coordinate variables $\mathcal{C}_{ba}, \mathcal{C}_{am}, \mathcal{C}_{ve}$.
\begin{equation}
V=sample\left( F_{cp},C \right), 
\end{equation}
where $sample\big( \cdot \big)$ refers to the process of using the coordinate vector $C$ ($\mathcal{C}_{ba}, \mathcal{C}_{am}, \mathcal{C}_{ve}$) to extract the corresponding coordinate features from $F_{cp}$, thereby forming the pixel value vectors $V$=($V_{ve}, V_{am}, V_{ba}$).

Although the categories of ambiguous pixels remain uncertain, they can be definitively categorized as either vessel or background pixels. Therefore, we employ the vessel and background vectors, $V_{ve}$ and $V_{ba}$, as driving engines. These engines interact, influence, and drive the ambiguous pixels from two perspectives, thereby bolstering the model's ability to effectively distinguish between different categories as follows:
\begin{equation}
\left\{ \begin{array}{l}
	    V_{v\mbox{-}dri}\left( i \right) =\sum_{j=1}^{L_{ve}}{\frac{\exp \left( V_{am}\left( i \right) ^T\cdot V_{ve}\left( j \right) \right)}{\sum{_{k=1}^{L_{ve}}\exp \left( V_{am}\left( i \right) ^T\cdot V_{ve}\left( k \right) \right)}}\cdot V_{ve}\left( j \right)}\\
	    V_{b\mbox{-}dri}\left( i \right) =\sum_{j=1}^{L_{ba}}{\frac{\exp \left( V_{am}\left( i \right) ^T\cdot V_{ba}\left( j \right) \right)}{\sum{_{k=1}^{L_{ba}}\exp \left( V_{am}\left( i \right) ^T\cdot V_{ba}\left( k \right) \right)}}\cdot V_{ba}\left( j \right)},\\
       \end{array} \right. 
\end{equation}
where $V_{am}\left( i \right)$ is the i-th channel of $V_{am}$, $V_{ve}\left( j \right)$ is the j-th channel of $V_{ve}$, $V_{ba}\left( j \right)$ is the j-th channel of $V_{ba}$. $V_{v\mbox{-}dri}\left( i \right)$ and $V_{b\mbox{-}dri}\left( i \right)$ are the i-th channels of $V_{v\mbox{-}dri}$ and $V_{b\mbox{-}dri}$. 
$L_{ve}$, $L_{am}$ and $L_{ba}$ represent the vector lengths of $V_{ve}$, $V_{am}$ and $V_{ba}$ respectively.
In this case, Vessel-Driven means that the ambiguous vector is regarded as Query, and the blood vessel vector is regarded as Key and Value. The Background-Driven is the same as above. 

As depicted on the right side of Fig.~\ref{fig: addr}, some ambiguous pixels have been successfully modified through the Vessel-Driven and Background-Driven processes. Subsequently, the two refined vectors $V_{v\mbox{-}dri}$ and $V_{b\mbox{-}dri}$ (where pixels identified as blood vessels in either vector are retained to enhance the representation of vessels, with all others remaining original ambiguous values) are combined with the definitive blood vessel and background pixels from $V_{ve}$ and $V_{ba}$. These are then compiled into a new vector, which is normalized to produce $\mathcal{W}_{dri}$ as the weighting factor.
Subsequently, the weight \(\mathcal{W}_{dri}\) is element-wise multiplied with the flattened \(F_{cp}\), followed by scanning through the Mamba block. The scanned result is then reshaped back into the feature map and $F_{cp}$ is added to the output as follows:
\begin{equation}
F_{addr}^{'}=Reshape\left(\mathcal{M}\left(\mathcal{W}_{dri}\cdot Flatten\left( F_{cp} \right) \right) \right) +F_{cp}.
\end{equation}

\begin{figure}[t]
    \centering
    \includegraphics[width=0.42\textwidth]{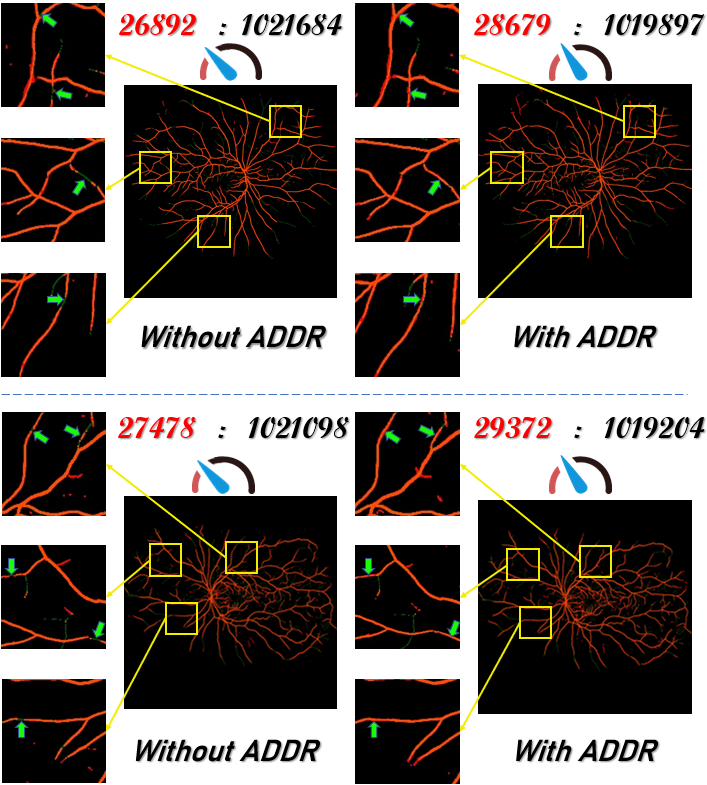}
    \vspace{-2mm}
    \caption{The figure shows the results of the ablation experiments with or without ADDR, where red represents the correct segmentation results, green represents the ground truth, and the arrows point to the noteworthy locations, please zoom in to check the details. Left: Results of the model without ADDR module. Right: Results of the model with ADDR module.}
    \label{fig:ADDRview}
    \vspace{-5mm}
\end{figure}

\subsection{Implementation Details and Evaluation Metrics} \label{IDEM}

\begin{table*}[!ht]
\centering
\caption{The performance comparison of our proposed Serp-Mamba against other state-of-the-art methods for vessel segmentation based on UWF-SLO images on three datasets.}
\vspace{-2mm}
\label{tab1}
\resizebox{\textwidth}{!}{%
\begin{tabular}{@{}c|c|c|cccccccc@{}}
\toprule \toprule
& & \multicolumn{2}{c|}{Dice $\uparrow$} & \multicolumn{2}{c|}{IoU $\uparrow$} & \multicolumn{2}{c|}{MCC $\uparrow$} & \multicolumn{2}{c}{BM $\uparrow$} \\ 

\cmidrule(l){3-10}  \multirow{-2}{*}{Dataset} & \multirow{-2}{*}{Model} & {\footnotesize Mean}  & \multicolumn{1}{c|}{\footnotesize Std}      
& {\footnotesize Mean}  & \multicolumn{1}{c|}{\footnotesize Std} 
& {\footnotesize Mean} & \multicolumn{1}{c|}{\footnotesize Std} 
& {\footnotesize Mean} & {\footnotesize Std} \\ 

\midrule    
& SCSNet\cite{wu2021scs} & 55.97   & \multicolumn{1}{c|}{\small {\color[HTML]{369DA2} \scriptsize $\pm2.85$}}  & 38.92 & \multicolumn{1}{c|}{\small {\color[HTML]{369DA2} \scriptsize $\pm2.75$}}   & 55.17  & \multicolumn{1}{c|}{\small {\color[HTML]{369DA2} \scriptsize $\pm2.79$}}   & 50.69    & \small {\color[HTML]{369DA2} \scriptsize $\pm3.37$}   \\
& SA-UNet\cite{guo2021sa} & 45.60   & \multicolumn{1}{c|}{\small {\color[HTML]{369DA2} \scriptsize $\pm20.13$}}  & 31.43 & \multicolumn{1}{c|}{\small {\color[HTML]{369DA2} \scriptsize $\pm14.54$}}   & 43.81  & \multicolumn{1}{c|}{\small {\color[HTML]{369DA2} \scriptsize $\pm22.09$}}   & 41.57    & \small {\color[HTML]{369DA2} \scriptsize $\pm21.65$}   \\
& DE-DCGCN-EE\cite{li2022dual} & 49.95   & \multicolumn{1}{c|}{\small {\color[HTML]{369DA2} \scriptsize $\pm2.71$}}  & 32.89 & \multicolumn{1}{c|}{\small {\color[HTML]{369DA2} \scriptsize $\pm2.41$}}   & 48.65  & \multicolumn{1}{c|}{\small {\color[HTML]{369DA2} \scriptsize $\pm2.67$}}   & 43.44    & \small {\color[HTML]{369DA2} \scriptsize $\pm2.76$}   \\
& CogSeg\cite{sang2021super} & 41.55   & \multicolumn{1}{c|}{\small {\color[HTML]{369DA2} \scriptsize $\pm2.36$}}  & 26.25 & \multicolumn{1}{c|}{\small {\color[HTML]{369DA2} \scriptsize $\pm1.87$}}   & 42.90  & \multicolumn{1}{c|}{\small {\color[HTML]{369DA2} \scriptsize $\pm2.06$}}   & 32.46    & \small {\color[HTML]{369DA2} \scriptsize $\pm4.99$}   \\
& SuperVessel\cite{hu2022supervessel} & 52.74   & \multicolumn{1}{c|}{\small {\color[HTML]{369DA2} \scriptsize $\pm4.47$}}  & 35.94 & \multicolumn{1}{c|}{\small {\color[HTML]{369DA2} \scriptsize $\pm4.11$}}   & 52.77  & \multicolumn{1}{c|}{\small {\color[HTML]{369DA2} \scriptsize $\pm4.36$}}   & 43.78    & \small {\color[HTML]{369DA2} \scriptsize $\pm4.63$}   \\
& SS-MAF\cite{zhang2022hard} & 57.86   & \multicolumn{1}{c|}{\small {\color[HTML]{369DA2} \scriptsize $\pm2.82$}}  & 40.77 & \multicolumn{1}{c|}{\small {\color[HTML]{369DA2} \scriptsize $\pm2.80$}}   & 57.45  & \multicolumn{1}{c|}{\small {\color[HTML]{369DA2} \scriptsize $\pm2.52$}}   & 50.78    & \small {\color[HTML]{369DA2} \scriptsize $\pm4.27$}   \\
& DS2F\cite{qiu2023rethinking} & 58.25   & \multicolumn{1}{c|}{\small {\color[HTML]{369DA2} \scriptsize $\pm3.40$}}  & 41.16 & \multicolumn{1}{c|}{\small {\color[HTML]{369DA2} \scriptsize $\pm3.41$}}   & 57.58  & \multicolumn{1}{c|}{\small {\color[HTML]{369DA2} \scriptsize $\pm3.38$}}   & 52.37    & \small {\color[HTML]{369DA2} \scriptsize $\pm4.12$}   \\
\cmidrule(l){2-10} & U-Mamba\cite{ma2024u} & 66.35 & \multicolumn{1}{c|}{\small {\color[HTML]{369DA2} \scriptsize $\pm2.42$}}  & 49.69 & \multicolumn{1}{c|}{\small {\color[HTML]{369DA2} \scriptsize $\pm2.68$}}   & 65.63  & \multicolumn{1}{c|}{\small {\color[HTML]{369DA2} \scriptsize $\pm2.26$}}   & 62.45    & \small {\color[HTML]{369DA2} \scriptsize $\pm3.80$}   \\ 
& VM-UNet\cite{ruan2024vm} & 64.91  & \multicolumn{1}{c|}{\small {\color[HTML]{369DA2} \scriptsize $\pm2.04$}}  & 48.09 & \multicolumn{1}{c|}{\small {\color[HTML]{369DA2} \scriptsize $\pm2.22$}}   & 64.16  & \multicolumn{1}{c|}{\small {\color[HTML]{369DA2} \scriptsize $\pm1.91$}}   & 60.91    & \small {\color[HTML]{369DA2} \scriptsize $\pm3.11$}   \\ 
& Mamba-UNet\cite{wang2402mamba} & 64.52  & \multicolumn{1}{c|}{\small {\color[HTML]{369DA2} \scriptsize $\pm2.41$}}  & 47.67 & \multicolumn{1}{c|}{\small {\color[HTML]{369DA2} \scriptsize $\pm2.59$}}   & 63.85  & \multicolumn{1}{c|}{\small {\color[HTML]{369DA2} \scriptsize $\pm2.28$}}   & 59.59    & \small {\color[HTML]{369DA2} \scriptsize $\pm3.40$}   \\ 
\multirow{-9}{*}{PRIME-FP20\cite{ding2020weakly}} & \textbf{Our Serp-Mamba} & \textbf{69.01}   & \multicolumn{1}{c|}{\small {\color[HTML]{369DA2} \scriptsize $\pm3.71$}}  & \textbf{52.93} & \multicolumn{1}{c|}{\small {\color[HTML]{369DA2} \scriptsize $\pm4.28$}}   & \textbf{68.41}  & \multicolumn{1}{c|}{\small {\color[HTML]{369DA2} \scriptsize $\pm3.68$}}   & \textbf{65.75}    & \small {\color[HTML]{369DA2} \scriptsize $\pm5.21$}   \\
\midrule
& SCSNet\cite{wu2021scs} & 57.45   & \multicolumn{1}{c|}{\small {\color[HTML]{369DA2} \scriptsize $\pm3.53$}}  & 40.39 & \multicolumn{1}{c|}{\small {\color[HTML]{369DA2} \scriptsize $\pm3.55$}}   & 56.76  & \multicolumn{1}{c|}{\small {\color[HTML]{369DA2} \scriptsize $\pm3.28$}}   & 52.28    & \small {\color[HTML]{369DA2} \scriptsize $\pm4.95$}   \\
& SA-UNet\cite{guo2021sa} & 52.94   & \multicolumn{1}{c|}{\small {\color[HTML]{369DA2} \scriptsize $\pm3.60$}}  & 36.08 & \multicolumn{1}{c|}{\small {\color[HTML]{369DA2} \scriptsize $\pm3.29$}}   & 52.00  & \multicolumn{1}{c|}{\small {\color[HTML]{369DA2} \scriptsize $\pm3.73$}}   & 58.24    & \small {\color[HTML]{369DA2} \scriptsize $\pm5.34$}   \\
& DE-DCGCN-EE\cite{li2022dual} & 56.39   & \multicolumn{1}{c|}{\small {\color[HTML]{369DA2} \scriptsize $\pm4.74$}}  & 39.42 & \multicolumn{1}{c|}{\small {\color[HTML]{369DA2} \scriptsize $\pm4.64$}}   & 56.14  & \multicolumn{1}{c|}{\small {\color[HTML]{369DA2} \scriptsize $\pm4.57$}}   & 48.89    & \small {\color[HTML]{369DA2} \scriptsize $\pm5.27$}   \\
& CogSeg\cite{sang2021super} & 53.35   & \multicolumn{1}{c|}{\small {\color[HTML]{369DA2} \scriptsize $\pm5.90$}}  & 36.60 & \multicolumn{1}{c|}{\small {\color[HTML]{369DA2} \scriptsize $\pm5.49$}}   & 52.69  & \multicolumn{1}{c|}{\small {\color[HTML]{369DA2} \scriptsize $\pm5.80$}}   & 47.62    & \small {\color[HTML]{369DA2} \scriptsize $\pm6.98$}   \\
& SuperVessel\cite{hu2022supervessel} & 57.05   & \multicolumn{1}{c|}{\small {\color[HTML]{369DA2} \scriptsize $\pm4.38$}}  & 40.04 & \multicolumn{1}{c|}{\small {\color[HTML]{369DA2} \scriptsize $\pm4.35$}}   & 56.28  & \multicolumn{1}{c|}{\small {\color[HTML]{369DA2} \scriptsize $\pm4.36$}}   & 52.47    & \small {\color[HTML]{369DA2} \scriptsize $\pm4.91$}   \\
& SS-MAF\cite{zhang2022hard} & 57.98   & \multicolumn{1}{c|}{\small {\color[HTML]{369DA2} \scriptsize $\pm4.49$}}  & 40.97 & \multicolumn{1}{c|}{\small {\color[HTML]{369DA2} \scriptsize $\pm4.52$}}   & 57.25  & \multicolumn{1}{c|}{\small {\color[HTML]{369DA2} \scriptsize $\pm4.36$}}   & 53.00    & \small {\color[HTML]{369DA2} \scriptsize $\pm5.28$}   \\
& DS2F\cite{qiu2023rethinking} & 57.16   & \multicolumn{1}{c|}{\small {\color[HTML]{369DA2} \scriptsize $\pm2.81$}}  & 40.07 & \multicolumn{1}{c|}{\small {\color[HTML]{369DA2} \scriptsize $\pm2.84$}}   & 56.39  & \multicolumn{1}{c|}{\small {\color[HTML]{369DA2} \scriptsize $\pm2.74$}}   & 51.72    & \small {\color[HTML]{369DA2} \scriptsize $\pm3.49$}   \\
\cmidrule(l){2-10} & U-Mamba\cite{ma2024u} & 57.59  & \multicolumn{1}{c|}{\small {\color[HTML]{369DA2} \scriptsize $\pm3.07$}}  & 40.51 & \multicolumn{1}{c|}{\small {\color[HTML]{369DA2} \scriptsize $\pm3.13$}}   & 56.72  & \multicolumn{1}{c|}{\small {\color[HTML]{369DA2} \scriptsize $\pm3.08$}}   & 53.77    & \small {\color[HTML]{369DA2} \scriptsize $\pm3.66$}   \\ 
& VM-UNet\cite{ruan2024vm} & 56.98  & \multicolumn{1}{c|}{\small {\color[HTML]{369DA2} \scriptsize $\pm3.12$}}  & 39.91 & \multicolumn{1}{c|}{\small {\color[HTML]{369DA2} \scriptsize $\pm3.13$}}   & 56.11  & \multicolumn{1}{c|}{\small {\color[HTML]{369DA2} \scriptsize $\pm2.97$}}   & 52.97    & \small {\color[HTML]{369DA2} \scriptsize $\pm4.50$}   \\ 
& Mamba-UNet\cite{wang2402mamba} & 56.07  & \multicolumn{1}{c|}{\small {\color[HTML]{369DA2} \scriptsize $\pm3.38$}}  & 39.04 & \multicolumn{1}{c|}{\small {\color[HTML]{369DA2} \scriptsize $\pm3.32$}}   & 55.28  & \multicolumn{1}{c|}{\small {\color[HTML]{369DA2} \scriptsize $\pm3.11$}}   & 51.48    & \small {\color[HTML]{369DA2} \scriptsize $\pm5.07$}   \\ 
\multirow{-9}{*}{MU-VS Center A\cite{wang2024advancing}} & \textbf{Our Serp-Mamba} & \textbf{61.06}   & \multicolumn{1}{c|}{\small {\color[HTML]{369DA2} \scriptsize $\pm1.62$}}  & \textbf{44.02} & \multicolumn{1}{c|}{\small {\color[HTML]{369DA2} \scriptsize $\pm1.67$}}   & \textbf{60.12}  & \multicolumn{1}{c|}{\small {\color[HTML]{369DA2} \scriptsize $\pm1.61$}}   & \textbf{58.37}    & \small {\color[HTML]{369DA2} \scriptsize $\pm1.41$}   \\
\midrule
& SCSNet\cite{wu2021scs} & 55.06   & \multicolumn{1}{c|}{\small {\color[HTML]{369DA2} \scriptsize $\pm3.13$}}  & 38.06 & \multicolumn{1}{c|}{\small {\color[HTML]{369DA2} \scriptsize $\pm2.99$}}   & 54.28  & \multicolumn{1}{c|}{\small {\color[HTML]{369DA2} \scriptsize $\pm3.05$}}   & 50.21    & \small {\color[HTML]{369DA2} \scriptsize $\pm4.16$}   \\
& SA-UNet\cite{guo2021sa} & 54.08   & \multicolumn{1}{c|}{\small {\color[HTML]{369DA2} \scriptsize $\pm3.02$}}  & 37.12 & \multicolumn{1}{c|}{\small {\color[HTML]{369DA2} \scriptsize $\pm2.85$}}   & 53.14  & \multicolumn{1}{c|}{\small {\color[HTML]{369DA2} \scriptsize $\pm3.08$}}   & 53.55    & \small {\color[HTML]{369DA2} \scriptsize $\pm4.79$}   \\
& DE-DCGCN-EE\cite{li2022dual} & 54.99   & \multicolumn{1}{c|}{\small {\color[HTML]{369DA2} \scriptsize $\pm3.22$}}  & 37.99 & \multicolumn{1}{c|}{\small {\color[HTML]{369DA2} \scriptsize $\pm3.05$}}   & 54.29  & \multicolumn{1}{c|}{\small {\color[HTML]{369DA2} \scriptsize \scriptsize $\pm3.22$}}   & 48.37    & \small {\color[HTML]{369DA2} \scriptsize $\pm3.27$}   \\
& CogSeg\cite{sang2021super} & 54.56   & \multicolumn{1}{c|}{\small {\color[HTML]{369DA2} \scriptsize $\pm2.82$}}  & 37.57 & \multicolumn{1}{c|}{\small {\color[HTML]{369DA2} \scriptsize $\pm2.66$}}   & 53.35  & \multicolumn{1}{c|}{\small {\color[HTML]{369DA2} \scriptsize $\pm2.64$}}   & 52.82    & \small {\color[HTML]{369DA2} \scriptsize $\pm3.29$}   \\
& SuperVessel\cite{hu2022supervessel} & 54.98   & \multicolumn{1}{c|}{\small {\color[HTML]{369DA2} \scriptsize $\pm3.08$}}  & 37.39 & \multicolumn{1}{c|}{\small {\color[HTML]{369DA2} \scriptsize $\pm2.89$}}   & 53.15  & \multicolumn{1}{c|}{\small {\color[HTML]{369DA2} \scriptsize $\pm2.96$}}   & 51.70    & \small {\color[HTML]{369DA2} \scriptsize $\pm3.20$}   \\
& SS-MAF\cite{zhang2022hard} & 56.03   & \multicolumn{1}{c|}{\small {\color[HTML]{369DA2} \scriptsize $\pm2.92$}}  & 38.97 & \multicolumn{1}{c|}{\small {\color[HTML]{369DA2} \scriptsize $\pm2.82$}}   & 54.91  & \multicolumn{1}{c|}{\small {\color[HTML]{369DA2} \scriptsize $\pm2.78$}}   & 53.10    & \small {\color[HTML]{369DA2} \scriptsize $\pm3.43$}   \\
& DS2F\cite{qiu2023rethinking} & 55.19   & \multicolumn{1}{c|}{\small {\color[HTML]{369DA2} \scriptsize $\pm2.60$}}  & 38.15 & \multicolumn{1}{c|}{\small {\color[HTML]{369DA2} \scriptsize $\pm2.48$}}   & 54.13  & \multicolumn{1}{c|}{\small {\color[HTML]{369DA2} \scriptsize $\pm2.42$}}   & 52.11    & \small {\color[HTML]{369DA2} \scriptsize $\pm3.47$}   \\
\cmidrule(l){2-10} & U-Mamba\cite{ma2024u} & 55.80  & \multicolumn{1}{c|}{\small {\color[HTML]{369DA2} \scriptsize $\pm2.97$}}  & 38.76 & \multicolumn{1}{c|}{\small {\color[HTML]{369DA2} \scriptsize $\pm2.86$}}   & 54.66  & \multicolumn{1}{c|}{\small {\color[HTML]{369DA2} \scriptsize $\pm2.82$}}   & 53.86    & \small {\color[HTML]{369DA2} \scriptsize $\pm3.45$}   \\ 
& VM-UNet\cite{ruan2024vm} & 55.42  & \multicolumn{1}{c|}{\small {\color[HTML]{369DA2} \scriptsize $\pm2.85$}}  & 38.39 & \multicolumn{1}{c|}{\small {\color[HTML]{369DA2} \scriptsize $\pm2.73$}}   & 54.31  & \multicolumn{1}{c|}{\small {\color[HTML]{369DA2} \scriptsize $\pm2.68$}}   & 52.88    & \small {\color[HTML]{369DA2} \scriptsize $\pm3.29$}   \\ 
& Mamba-UNet\cite{wang2402mamba} & 55.64  & \multicolumn{1}{c|}{\small {\color[HTML]{369DA2} \scriptsize $\pm3.04$}}  & 38.60 & \multicolumn{1}{c|}{\small {\color[HTML]{369DA2} \scriptsize $\pm2.92$}}   & 54.58  & \multicolumn{1}{c|}{\small {\color[HTML]{369DA2} \scriptsize $\pm2.92$}}   & 52.17    & \small {\color[HTML]{369DA2} \scriptsize $\pm3.18$}   \\ 
\multirow{-9}{*}{MU-VS Center B\cite{wang2024advancing}} & \textbf{Our Serp-Mamba} & \textbf{57.10}   & \multicolumn{1}{c|}{\small {\color[HTML]{369DA2} \scriptsize $\pm3.96$}}  & \textbf{40.06} & \multicolumn{1}{c|}{\small {\color[HTML]{369DA2} \scriptsize $\pm3.83$}}   & \textbf{56.34}  & \multicolumn{1}{c|}{\small {\color[HTML]{369DA2} \scriptsize $\pm3.57$}}   & \textbf{53.92}    & \small {\color[HTML]{369DA2} \scriptsize $\pm6.90$}  \\
\bottomrule \bottomrule
\end{tabular}}
\label{tab11}
\vspace{-5mm}
\end{table*}

We use an NVIDIA Tesla V100 GPU with 32GB of memory for all experiments. To ensure fairness in the experiments, we resize all inputs to 1024×1024. During training, we use the Adam optimizer with an initial learning rate of 0.0001 and weight decay of 0.0001. All models are trained for 12,000 iterations.
Following previous works \cite{qiu2023rethinking,wang2024advancing}, to make a comprehensive quantitative assessment of these segmentation models, we report the Dice coefficient (Dice), Intersection over Union (IoU), Matthews correlation coefficient (MCC) and Bookmaker informedness (BM) metrics. Meanwhile, we conduct a five-fold cross-validation experiment to enhance the robustness and reliability of our findings.

\begin{table}[t]
\centering
\caption{The categories and image resolution of the datasets. }
\vspace{-2mm}
\label{tab1}
\resizebox{\columnwidth}{!}{%
\begin{tabular}{@{}c|m{6cm}|c@{}}  % 使用 m{6cm} 来替代 p{6cm} 实现垂直居中
\toprule
Dataset & Categories & Resolution \\
\midrule
PRIME-FP20 & Diabetic Retinopathy & 4000×4000 \\
\midrule
MU-VS Center A & Normal, Retinal Vein Occlusion & 3900×3072 \\
\midrule
MU-VS Center B & Diabetic Retinopathy, Retinal Vein Occlusion, Retinitis Pigmentosa, Retinal Artery Occlusion, Central Serous Chorioretinopathy & 3900×3072\\
\bottomrule
\end{tabular}%
}
\label{datasets}
\vspace{-5mm}
\end{table}

\begin{table*}[t]
\centering
\caption{Ablation study of the proposed method (including SIA and ADDR) on three datasets.}
\vspace{-2mm}
\label{tab1}
\resizebox{\textwidth}{!}{%
\begin{tabular}{@{}c|c|cc|cc|cc|cc|cc@{}}
\toprule \toprule
\multirow{2}{*}{Datasets} & \multirow{2}{*}{Methods} & \multicolumn{2}{c|}{Modules} &  \multicolumn{2}{c|}{Dice $\uparrow$} & \multicolumn{2}{c|}{IoU $\uparrow$} & \multicolumn{2}{c|}{MCC $\uparrow$} & \multicolumn{2}{c}{BM $\uparrow$} \\
& & SIA & ADDR & Mean & Std & Mean & Std & Mean & Std & Mean & Std \\ 
\midrule    
\multirow{4}{*}{PRIME-FP20\cite{ding2020weakly}} & Baseline & & & 66.35 & $\pm2.42$ & 49.69 & $\pm2.68$ & 65.63 & $\pm2.26$ & 62.45 & $\pm3.80$ \\
 & M1 & \Checkmark & & 68.37 & $\pm4.60$ & 52.13 & $\pm5.19$ & 67.78 & $\pm4.49$ & 64.06 & $\pm5.25$ \\
 & M2 & & \Checkmark & 67.79 & $\pm2.32$ & 51.32 & $\pm2.63$ & 67.02 & $\pm2.19$ & 65.18 & $\pm3.55$ \\
\rowcolor{gray!15}
 & \textbf{Ours} & \Checkmark & \Checkmark & \textbf{69.01} & $\pm3.40$ & \textbf{52.93} & $\pm3.41$ & \textbf{68.41} & $\pm3.38$ & \textbf{65.75} & $\pm4.12$ \\
\midrule
\multirow{4}{*}{MU-VS Center A\cite{wang2024advancing}} & Baseline & & & 57.59 & $\pm3.07$ & 40.51 & $\pm3.13$ & 56.72 & $\pm3.08$ & 53.77 & $\pm3.66$ \\
 & M1 & \Checkmark & & 59.51 & $\pm3.00$ & 42.42 & $\pm3.11$ & 58.59 & $\pm2.87$ & 56.94 & $\pm5.08$ \\
 & M2 & & \Checkmark & 58.44 & $\pm3.09$ & 41.35 & $\pm3.16$ & 57.63 & $\pm2.89$ & 54.40 & $\pm5.03$ \\
\rowcolor{gray!15}
 & \textbf{Ours} & \Checkmark & \Checkmark & \textbf{61.06} & $\pm1.62$ & \textbf{44.02} & $\pm1.67$ & \textbf{60.12} & $\pm1.61$ & \textbf{58.37} & $\pm1.41$ \\
\midrule
\multirow{4}{*}{MU-VS Center B\cite{wang2024advancing}} & Baseline & & & 55.80 & $\pm2.97$ & 38.76 & $\pm2.86$ & 54.66 & $\pm2.82$ & 53.86 & $\pm3.45$ \\
 & M1 & \Checkmark & & 55.99 & $\pm2.99$ & 38.93 & $\pm2.88$ & 54.90 & $\pm2.86$ & 53.80 & $\pm3.43$ \\
 & M2 & & \Checkmark & 56.02 & $\pm3.63$ & 38.99 & $\pm3.49$ & 55.12 & $\pm3.54$ & 53.90 & $\pm4.47$ \\
\rowcolor{gray!15}
 & \textbf{Ours} & \Checkmark & \Checkmark & \textbf{57.10} & $\pm3.96$ & \textbf{40.06} & $\pm3.83$ & \textbf{56.30} & $\pm3.57$ & \textbf{53.92} & $\pm6.90$ \\
\bottomrule \bottomrule
\end{tabular}}
\vspace{-4mm}
\end{table*}

\section{Experiments and Results}
% For this section, first, we show the performance of our method on three different datasets. Also, we compare other novel super-resolution segmentation and Mamba-based methods with our method. In addition, we conducted abundant ablation experiments to prove the effectiveness of our method.

\subsection{Datasets}
To validate the efficacy of our approach, we carry out experiments on three publicly available UWF-SLO datasets, detailed as follows:
\noindent \textbf{PRIME-FP20 dataset \cite{ding2020weakly}}: The PRIME-FP20 dataset includes 15 UWF-SLO images, which were captured using Optos California and 200Tx cameras (Optos plc, Dunfermline, UK) \cite{kato2019quantitative}. Each image is accompanied by a binary mask delineating its vessels. We preprocess the images following previous work \cite{qiu2023rethinking}, using the official masks to exclude invalid regions.
\noindent \textbf{MU-VS Center A and B \cite{wang2024advancing}}:
The MU-VS Center A and B datasets each contain 30 UWF-SLO images from two different medical centers, using Optos California and 200Tx cameras (Optos plc, Dunfermline, UK) \cite{kato2019quantitative} for capture. Each center includes images from patients with different disease categories (shown in Table~\ref{datasets}).

\subsection{Comparison with State-of-the-art Methods}
\begin{figure*}[t]
    \centering
\includegraphics[width=1\textwidth]{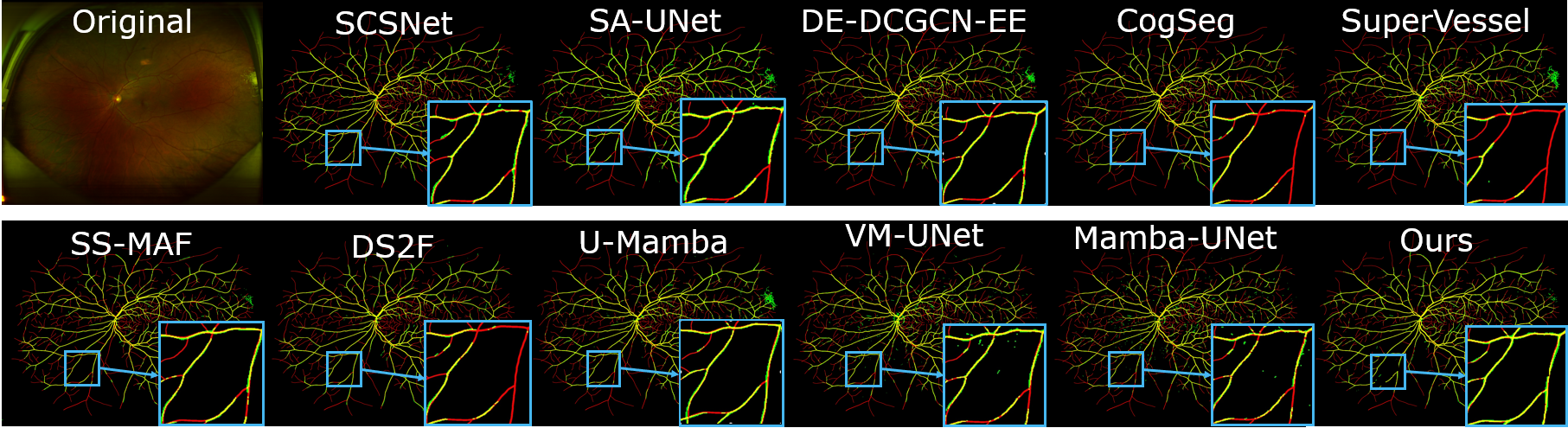}
    \vspace{-6mm}
    \caption{The visualization results for all models are displayed for the dataset MU-VS Center A, as an example. In the figure, red represents the ground truth, while yellow indicates the correct segmentation results that overlap with the ground truth. Please zoom in for the best view.}
    \label{fig:output}
    \vspace{-4mm}
\end{figure*}

We compare our method against a range of state-of-the-art methods as detailed in Table \ref{tab11}. To ensure a comprehensive and holistic comparison, these methods include models specifically designed for \textbf{vessel segmentation} (including SCSNet\cite{wu2021scs}, SA-UNet\cite{guo2021sa}, DE-DCGCN-EE\cite{li2022dual}) and those optimized for \textbf{high-resolution segmentation} (including CogSeg\cite{sang2021super}, SuperVessel\cite{hu2022supervessel}, SS-MAF\cite{zhang2022hard}, DS2F\cite{qiu2023rethinking}). Meanwhile, we include comparisons with several \textbf{Mamba-based segmentation} methods (including U-Mamba\cite{ma2024u}, VM-UNet\cite{ruan2024vm}, Mamba-UNet\cite{wang2402mamba}).
The experimental results presented in Table \ref{tab11} indicate that our Serp-Mamba method achieves the best performance across four metrics on three different datasets. Visual comparisons are illustrated in Fig.~\ref{fig:output}.

\begin{figure}[t]
    \centering
\includegraphics[width=0.42\textwidth]{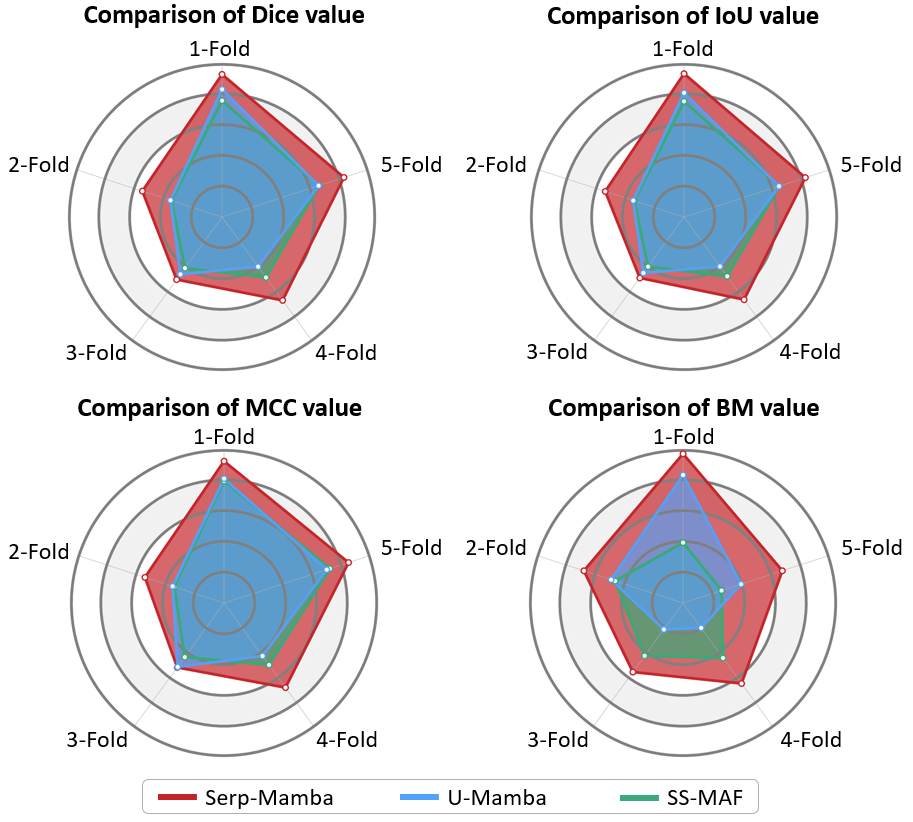}
    \vspace{-3mm}
    \caption{Radar chart illustrating the performance of various models across five folds of the MU-VS Center A dataset, particularly focusing on our method and other two top-performing compared models, U-Mamba\cite{ma2024u} and SS-MAF\cite{zhang2022hard}. }
    \label{fig:fivefold analysis}
    \vspace{-4mm}
\end{figure}

The experimental results on the PRIME-FP20 dataset show that Mamba-based methods (with Dice scores above 60\%) slightly outperform methods that do not incorporate Mamba. On the other hand, both types of methods achieve comparable performance on the MU-VS Center A and MU-VS Center B datasets. Specifically, in Center A, the Dice scores are around 57\%, and the IOU scores are about 40\%, whereas in Center B, the Dice scores are approximately 55\% and IOU scores around 38\%. These results reveal that models tend to perform slightly weaker in Center B than in Center A, which may be attributed to the increased complexity and diversity of disease categories present in Center B.

It’s worth noting that while the aforementioned Mamba-based methods have demonstrated commendable results, they are primarily designed for general medical image segmentation. However, our Serp-Mamba is the first network specifically tailored for high-resolution UWF-SLO vessel segmentation. Owing to its effective capture of tubular vascular features and dual refinement against the vascular and background, our method has achieved significant performance improvements across all three datasets. For instance, on the PRIME-FP20 dataset, Serp-Mamba reaches a Dice score of 69.01\%, an IOU of 52.93\%, an MCC of 68.41\%, and a BM of 65.75\%.
Additionally, the five-fold experimental results for the MU-VS Center A dataset are displayed in Fig.~\ref{fig:fivefold analysis}. The radar charts illustrate that our method surpasses the comparative methods across multiple dimensions and consistently outperforms others in each of the five-fold cross-validations, highlighting the stable performance advantage of our approach.

\subsection{Ablation Studies}
\label{ablation}
To confirm the effectiveness of our proposed method, we perform ablation studies (shown in Table \ref{tab22}) for SIA and ADDR on the PRIME-FP20, MU-VS Center A, and MU-VS Center B datasets, respectively. 
Specifically, we test the following network configurations: (i) Baseline: we employ the classical U-shaped U-Mamba backbone network as the baseline model; (ii) M1: we include the SIA scan mechanism, and without the ADDR module, to test whether SIA enables the Mamba block to improve the accuracy of segmentation after scanning along the learnable vessel path; (iii) M2: we remove the SIA scan mechanism and incorporate the ADDR module based on M1 to verify the effect of ADDR on addressing the category imbalance problem and enhancing vessel continuity; (iv) Ours: our Serp-Mamba includes both the SIA scan mechanism and the ADDR module.   

\begin{figure}[t]
    \centering
\includegraphics[width=0.46\textwidth]{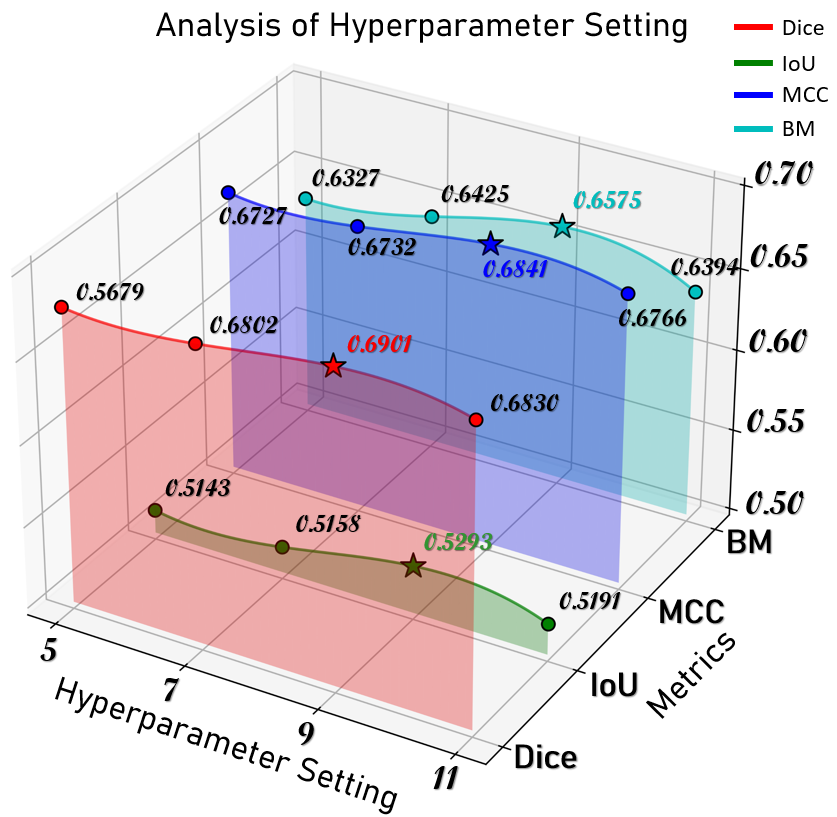}
    \vspace{-3mm}
    \caption{Ablation experiment results on serpentine scan path length $l$. The best result for each metric is marked with a star \mystar.}
    \label{fig:hype-analysis}
    \vspace{-6mm}
\end{figure}

The experimental results in Table~\ref{tab22} reveal enhancements for both models M1 and M2 across all four metrics compared to the Baseline model. For example, M1 exhibited an improvement of 2.02\% in Dice, 2.44\% in IoU, 2.15\% in MCC, and 1.61\% in BM on PRIME-FP20. These enhancements suggest that the learnable vessel scanning paths, guided by the SIA, effectively focused the Mamba's attention on the critical vessel features.
The metrics for M2 also exhibit improvement compared to the Baseline. To intuitively assess the impact of ADDR on segmentation, particularly regarding category imbalance issues and vessel continuity, we compare the results of M1 (without ADDR) and Serp-Mamba (with ADDR), presented in Fig.~\ref{fig:ADDRview}. The comparison reveals that ADDR not only improves the segmentation of previously disconnected vessel regions but also better balances the proportion of vessels to background pixels, effectively addressing the class imbalance problem. Our Serp-Mamba achieves the best results, demonstrating that the combination of SIA and ADDR can synergistically enhance model performance.
To assess the impact of the serpentine scanning path length $l$ on the performance, we conduct hyperparameter ablation experiments with path lengths of 5, 7, 9, and 11. The experimental data reveals a performance increase as the path length extended from 5 to 9, with peak performance at a length of 9, followed by a decline when the path length increased to 11. Based on these findings, a path length of 9 was chosen as optimal. To visually illustrate these trends, we plot a 3D line graph depicting the relationship between path length and performance in Fig.~\ref{fig:hype-analysis}.

\section{Disccusion}
UWF-SLO vessel segmentation is crucial in ophthalmology. It enables precise identification of retinal vessel morphology and abnormalities, aiding in the detection and monitoring of conditions like diabetic retinopathy and glaucoma, thus improving diagnostic accuracy and treatment outcomes \cite{tang2024applications}.
However, UWF-SLO images are high-resolution and feature complex vessel structures, making segmentation challenging. The small and tortuous nature of the vessels adds to this complexity. Additionally, the high resolution of these images intensifies the issue of category imbalance.

The SSM recently brought by Mamba has been extensively demonstrated to possess powerful long-range modeling capability. The high resolution of UWF-SLO images, which provides more information, also raises the challenge of accurate segmentation. Hence, we propose a novel Serp-Mamba to segment vessels in UWF-SLO images in this paper.

Inspired by the intricate and winding structure of retinal vessels, we devise the Mamba block to navigate learnable vascular paths in a manner like a snake’s curving movement, termed the Serpentine Interwoven Adaptive (SIA) scan mechanism, which enhances vessel continuity. By incorporating steps such as centroid sampling, deformation, straightening, and concatenation, the Mamba block focuses more acutely on delicate vascular features, thereby preserving vascular connectivity.
We also recognize that the category imbalance is particularly severe in high-resolution vascular images from UWF-SLO (only 2.7\% in Fig.~\ref{fig:comparison} (a)). To effectively address this challenge, we develop the Ambiguity-Driven Dual Recalibration (ADDR) module. By employing dual-driven branching for blood vessels and background pixels, this module aids the model in more accurately distinguishing the correct categories by enhancing its focus on vessel features. Meanwhile, we add the Continuity Perception strategy to the ADDR, which further enhances the continuity of the blood vessels.
Compared with other state-of-the-art high-resolution and Mamba-based segmentation methods, our approach delivers superior accuracy in segmenting vessels in UWF-SLO images. As detailed in Table \ref{tab11}, our method achieves a Dice coefficient of 69.01\% (best results) on the PRIME-FP20 dataset. Moreover, it continues to outperform compared methods on the MU-VS Center A and B datasets.

%Limitations 动态选择扫描路径长度和形态
A promising avenue for enhancing our methodology is to further refine the SIA scan mechanism, which currently employs a dynamically deformable scanning path to closely align with the fundus blood vessels, allowing the Mamba block to concentrate on vascular features. However, this approach may also have its limitations, notably the varying widths of each blood vessel, which may suggest the need for a more tailored scanning path size. A potential solution could involve designing a method that dynamically selects the width and length of the serpentine scanning path to better match the individual characteristics of each vessel.

\section{Conclusion}
In this paper, we present Serp-Mamba, the first Mamba-based model specifically designed for vessel segmentation in high-resolution UWF-SLO images. Acknowledging the unique characteristics of vascular curvature, we devise a Serpentine Interwoven Adaptive (SIA) scan mechanism. This design enables the Mamba's scanning path to adapt and align with vessel tracks, enhancing segmentation accuracy and improving vessel continuity.
Furthermore, we introduce a novel Ambiguity-Driven Dual Recalibration (ADDR) module that refines visual features through the dual-driven of vessel and background, effectively addressing the category imbalance problem. Experiments on three publicly available datasets demonstrate that our method outperforms other state-of-the-art models.
%%
% Future work includes exploring the introduction of prior knowledge and more complex topological principles.

\bibliographystyle{IEEEtran}
\bibliography{arxiv}

\end{document}